\documentclass[conference]{IEEEtran}
\IEEEoverridecommandlockouts
\usepackage{cite}
\usepackage{amsmath,amssymb,amsfonts}
\usepackage{algorithmic}
\usepackage{graphicx}
\usepackage{subcaption}
\usepackage{textcomp}
\usepackage[export]{adjustbox}
\usepackage{xcolor}
\def\BibTeX{{\rm B\kern-.05em{\sc i\kern-.025em b}\kern-.08em
    T\kern-.1667em\lower.7ex\hbox{E}\kern-.125emX}}
\begin{document}

\title{ Weakly-supervised fire segmentation by visualizing intermediate CNN layers 
}

\author{\IEEEauthorblockN{1\textsuperscript{st} Milad Niknejad}
\IEEEauthorblockA{\textit{Instituto de Sistemas e Robotica, } \\
\textit{ Instituto Superior Tecnico, University of Lisbon}\\
Lisbon, Portugal \\
milad3n@gmail.com}
\and
\IEEEauthorblockN{2\textsuperscript{nd} Alexandre Bernardino}
\IEEEauthorblockA{\textit{Instituto de Sistemas e Robotica, } \\
\textit{ Instituto Superior Tecnico, University of Lisbon}\\
Lisbon, Portugal \\
alex@isr.tecnico.ulisboa.pt
}

}

\maketitle

\begin{abstract}
Fire localization in images and videos is an important step for an autonomous system to combat fire incidents. State-of-art image segmentation methods based on deep neural networks require a large number of pixel-annotated samples to train Convolutional Neural Networks (CNNs) in a fully-supervised manner. In this paper, we consider weakly supervised segmentation of fire in images, in which only image labels are used to train the network. We show that in the case of fire segmentation, which is a binary segmentation problem, the mean value of features in a mid-layer  of classification CNN  can perform better than conventional Class Activation Mapping (CAM) method. We also propose to further improve the segmentation accuracy by adding a rotation equivariant regularization loss on the features of the last convolutional layer. Our results show noticeable improvements over baseline method for weakly-supervised fire segmentation.
\end{abstract}


\section{Introduction}
Artificial intelligence can help to prevent fire incidents by early detection of fire and smoke, and identifying the incident location. Fire localization in images and videos is the first step for an autonomous system to prevent the fire incidents. In this paper, we address the problem of weakly supervised fire localization, using only image level labels. The localization has pixel-level accuracy which outputs a segmentation mask. Compared to bounding boxes, pixel-wise segmentation achieve more precise localization accuracy which is useful in applications such as georefrencing fire from arial vehicles, and feeding into fire propagation models. 

Like other areas in computer vision, state-of-the-art results for fire detection  have been achieved through using Convolutional Neural Networks (CNNs). However, training CNNs for image segmentation requires a large number of annotated images. Annotating these images is time-consuming and requires a lot of effort. Moreover, there are some ambiguities in annotating the pixels especially in the boundaries of the objects like fire and smoke. To avoid these issues, weakly-supervised methods have been proposed for segmentation with less expensive annotation \cite{vernaza2017learning, wei2018revisiting, pathak2015constrained}. Instead of pixel level annotation, these methods use image-level or bounding box annotations in order to infer the segmentation mask. 

State-of-the-art weakly supervised methods are mainly based on visualization of the networks trained for image recognition using Class Activation Mapping (CAM) \cite{zhou2016learning}.
CAM uses a weighted average of the activations in the last convolutional layer to obtain the segmented masks. The masks obtained by CAM can then be used as the ground truth for training a separate fully convolutional  network (FCN) to produce  improved  segmentation masks \cite{wei2018revisiting}.
However, it is well-known that the visualized masks obtained by CAM focuses on the discriminative part of objects, but can be improved for the localization tasks \cite{choe2020evaluating,bae2020rethinking,wei2018revisiting}. 
It is also known that using intermediate features in a CNN has more accurate localization information, though is less accurate for classification of pixels. So, many image segmentation methods such as U-net \cite{ronneberger2015u}, use skip connections between the encoder and decoder layers of CNNs to compensate for the loss in the localization accuracy.

In this paper, we propose a weakly-supervised method for fire segmentation in images. We show that in the case of binary fire segmentation, the CAM method covers most part of the object and does not focus on discriminative parts of fire. We propose to use the intermediate layers of CNNs, which contain more accurate spatial information, to further improve the initial masks obtained by CAM.  Our experiments show that in the case of binary object segmentation of fire images, using the activations in the intermediate  layers of CNNs leads to more accurate initial segmented masks. These masks allow obtaining more accurate segmentation results after training the second network via pixel supervision.

In the following, sections, we first mention related works of weakly-supervised segmentation, and fire detection methods. The proposed method is then described in detail. Finally, our proposed method is compared to other state-of-the-art methods in the experimental results section.

\section{Related works}
Many works have considered detection of fire in images. Traditional methods typically used hand-crafted features mainly based on colors \cite{celik2009fire}, \cite{chen2004early}. More recent methods use the features obtained by CNNs \cite{dunnings2018experimentally, barmpoutis2019fire}. Beyond fire recognition, some works have considered the fire localization based on bounding boxes \cite{zhang2016deep, chaoxia2020information} or segmentation masks \cite{harkat2021fire, frizzi2021convolutional}. In \cite{harkat2021fire}, a deeplan v3+ \cite{chen2018encoder} is adapted for fire segmentation. \cite{frizzi2021convolutional} proposes a new CNN architecture for fire segmentation.  Pixel localization can be used as an input for fire propagation models, while bounding boxes, does not have such precision. It also lead to more precise geo-referencing especially for aerial images. All mentioned methods for fire localization are based on fully supervised approaches. They require pixel-wise or bounding box labeling which are time-consuming to obtain. Even training on existing annotated fire datasets may not work in other environments such as different types of forest vegetation, different seasons, or different distances of aerial vehicles that capture the images. They require new pixel annotations in the new environment or use domain adaptation methods.

Semantic segmentation is the task of assigning a label from a predefined category to each pixel in the image. Like many other areas in computer vision, in semantic segmentation, state-of-the-art results are achieved using CNNs \cite{chen2017deeplab, long2015fully}. However, fully-supervised segmentation requires a large number of annotated images, which are time-consuming and expensive to obtain. To solve this issue, many methods consider a weakly supervised approach in which less expensive annotations (e.g. image labels) are used to infer the segmentation masks \cite{wei2018revisiting, pathak2015constrained}. Most weakly supervised methods include CAM \cite{zhou2016learning} in their methods to obtain the initial segment cues \cite{vernaza2017learning, wei2018revisiting}. CAM uses a weighted average of the feature maps (activations) in the last convolution layer in the CNN trained for classification to produce class-specific localization. The weights are proportional to the fully connected layer weights, which determines the importance of the feature maps to a specific class. To obtain the initial masks, the segmentation methods threshold the CAM output proportional to its maximum value. The masks obtained by CAM often focus on small discriminative parts of objects, and do not cover the entire objects. For example in the case of bird classification, the localization map may only cover the head of the bird. \cite{huang2018weakly} proposed a method in which the CAM cues are iteratively expanded by  seeded region growing. \cite{wei2017object} uses an iterative adversarial erasing method in which the discriminative parts are iteratively erased to obtain the complements parts of the objects. All mentioned works consider multi-class weakly supervised problems. In this paper we consider the binary case of detecting fire in images.

We found that in the fire binary classification, unlike general multi-label classification, masks obtained by CAM covers a large part of fire in images. This can be observed in examples in Fig. \ref{fig:segmented1}. 
However, as it can be seen in Fig \ref{fig:segmented1}, CAM mask has a bulb like shape for fire and does not correspond to the details of the original mask.

\subsection{Class activation mapping}
\label{sec:cam}
Here, we describe the formulation of CAM and an equivalent formulation that will allow us to derive our proposed approach.  Let $\mathbf{A}^{k} \in \mathbb{R}^{H \times W}$ be the $k^{th}$ feature map with the spatial resolution $H \times W$ in the last convolution layer of the classification CNN, and $\mathbf{A}^{k}_{i,j}$  denote the entry in $i$ and $j$ position. The output score for the class $c$ before the soft-max is computed by
\begin{equation}
\label{eq:cam1}
{y}_c=\sum_k w_k^c \overbrace{\frac{1}{Z} \sum_i \sum_j}^\text{global average} \underbrace{\mathbf{A}_{ij}^{k}}_{features}
\end{equation}
where $Z=H \times W$, and $w_k^c$ measures the importance of the $k^{th}$ feature in the $c^{th}$ class, which is computed as the weight that connects the global averaged pool of the $k^{th}$ feature to the class score in the fully connected layer of the classification network. 
The network is trained by the cross entropy loss between one-hot ground truth label vector $\mathbf{y}$, and the above output after applying a (soft-max or sigmoid) function.

\begin{figure*}
 \begin{subfigure}{.49\textwidth}
\includegraphics[width=\textwidth, frame]{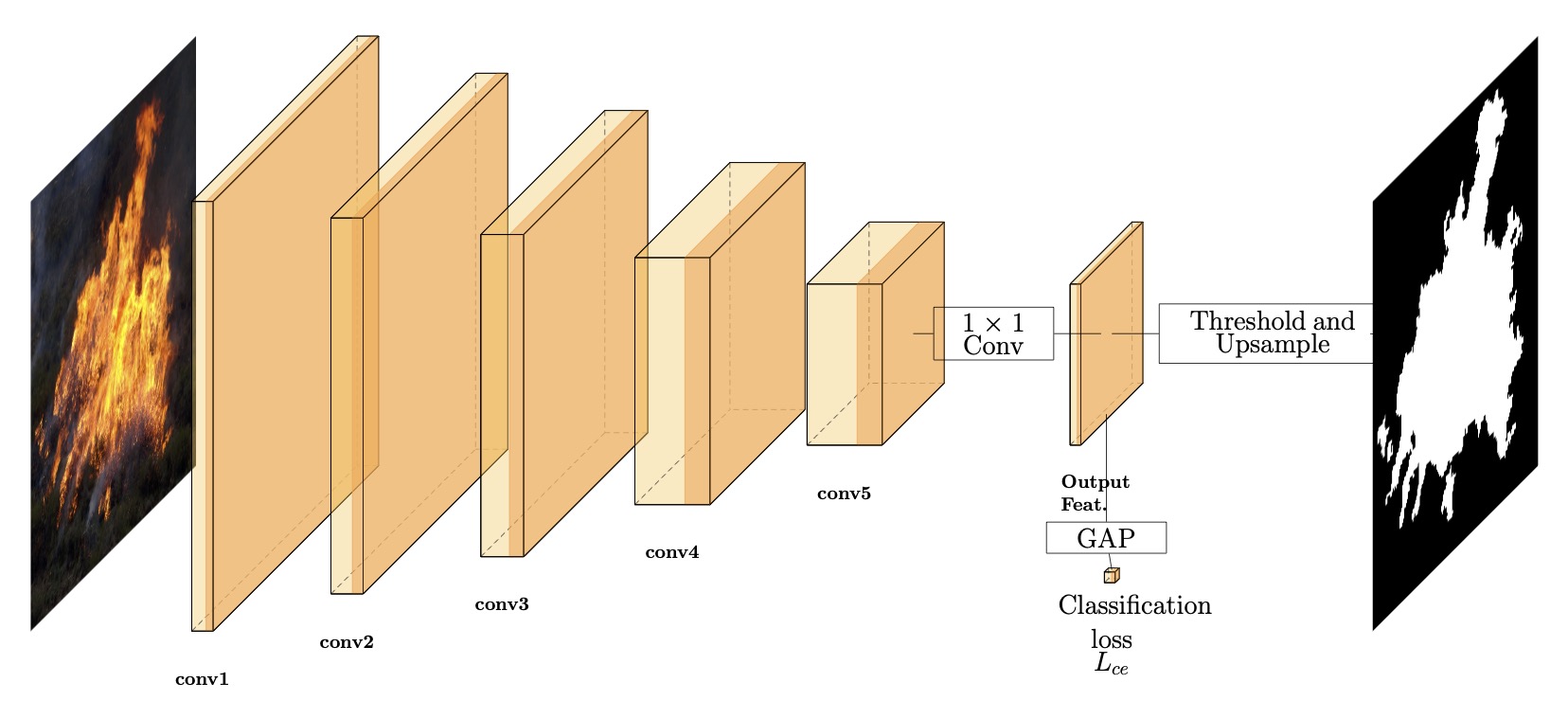} \caption{}
\end{subfigure} \hfill
\begin{subfigure}{.49\textwidth}
\includegraphics[width=\textwidth, frame]{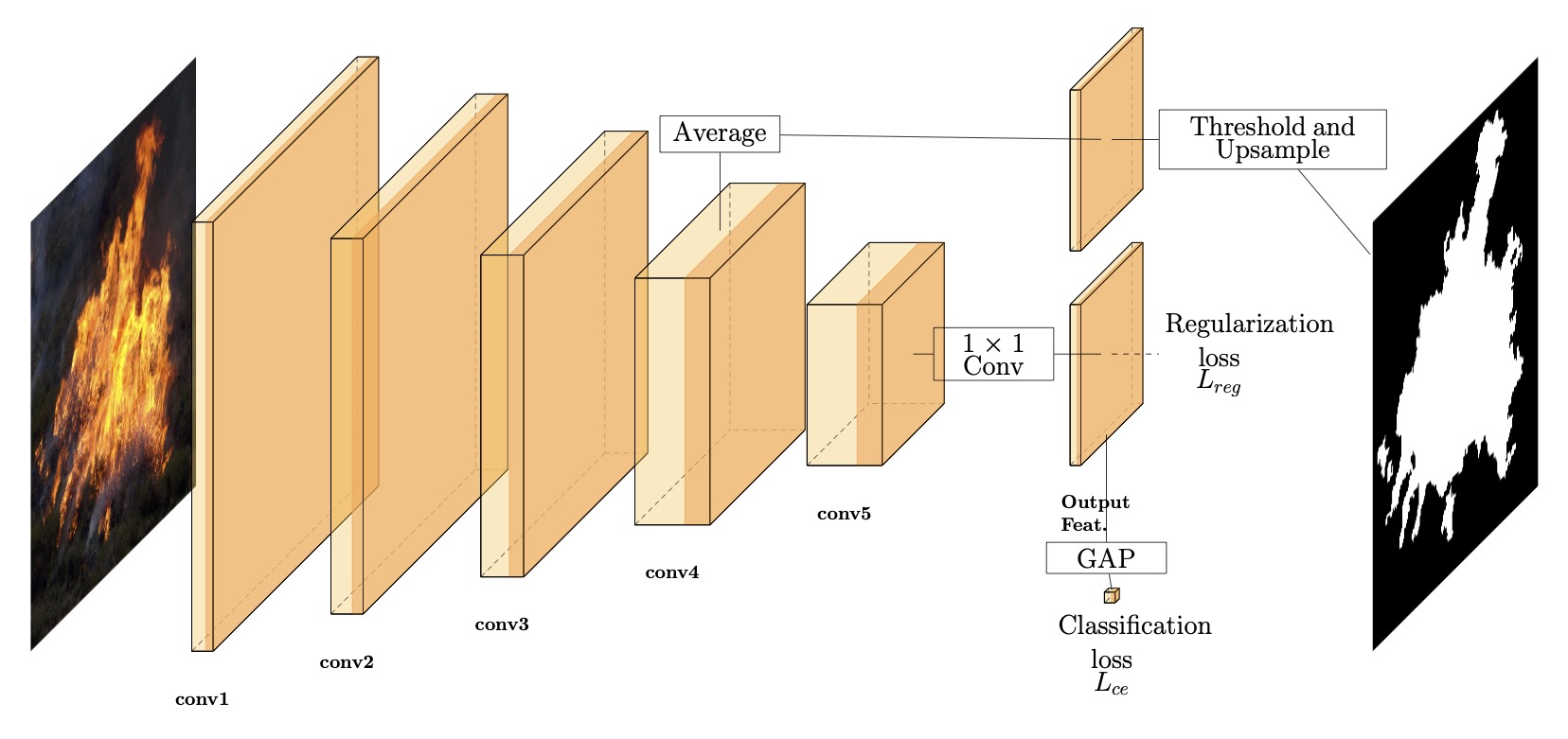} \caption{}
\end{subfigure}
\caption{The architecture of the networks for obtaining the initial masks (a) the equivalent CAM  after rearranging the formulation of the fully connected layer and (b) proposed method}
\label{fig:network}
\end{figure*}

By rearranging the sums in equation \ref{eq:cam1}, and since $Z$ is constant, the above equation can be written as 
\begin{equation}
y_c=\underbrace{\frac{1}{Z}\sum_i \sum_j}_{GAP} \overbrace{\sum_k w_k^c \mathbf{A}_{ij}^{k}}^\text{$1 \times 1$ Conv}.
\end{equation}
The above equation suggests an equivalent classification network in which $1 \times 1$ convolution is applied to the convolutional layer before pooling and the pooling is applied to the resulting features to get the classification scores.  The masks can be obtained by thresholding the features of the last layer. This formulation was also understood in \cite{choe2020evaluating}. The architecture of the network is shown in Fig. \ref{fig:network} (a). Note that this new architecture is the base for our proposed method.

\section{Proposed method}
\subsection{Mid-layer visualization}
For  fire segmentation, we propose to use mid-layer activations of a CNN to obtain the initial segmented mask. Generally, features from the early layers of a CNN have more localization accuracies, since some localization information is lost due to sequential pooling, while the features in deeper layers are better for classification. This trade-off has been noticed in some semantic segmentation methods, and is one of the motivations for  skip connections in those methods \cite{ronneberger2015u}. Unlike CAM which uses the last convolutional layer for obtaining the mask, we use the penultimate convolutional layer.  Let $\mathbf{B}  \in \mathbb{R}^{H_m \times W_m \times K}$ be the  penultimate convolution layer of the CNN  in the equivalent network  (in Fig. \ref{fig:network} (a)) and $\mathbf{B}^{k} \in \mathbb{R}^{H_m \times W_m}$ be its $k^{th}$ feature map . We obtain the activation map by averaging  the feature maps at each position $i,j$ in penultimate convolution layer as
\begin{equation}
\mathbf{M}_{i,j}=\frac{1}{K} \sum_k \mathbf{B}^{k}_{i,j}
\end{equation}
where $K$ is the number of features in the penultimate layer (see Fig. \ref{fig:network} (b)).
We found that this approach performs better for binary fire class compared to CAM. One reason may be the better localization accuracy for the earlier layers of the CNN as mentioned above. Our problem is to obtain binary masks, while CAM aims to produce general multi-class masks. In the binary case, the problem is equivalent to salient object segmentation or foreground segmentation problem. The average of the mid-layer activations has already been shown to have promising performance to obtain foreground segmentation in \cite{saleh2017incorporating}.

The masks are then obtained by upsampling $\mathbf{M}$ to the image dimensions, and setting the values below a threshold to zero, and the remaining to one. The threshold value, similar to CAM, is obtained by a factor of the maximum value of the output i.e. $ \tau  \max \mathbf{M}$, where $\tau$ is a constant.

\subsection{Consistency regularization}
Weakly-supervised segmentation methods based on CAM are trained merely on image label classification losses. However, their performance is measured on the pixel-level losses. Here, inspired by semi-supervised methods, we propose to add a consistency regularization loss which is dependent on the output pixels. Consistency regularization methods have been extensively used in the semi-supervised image classification methods \cite{sohn2020fixmatch, sajjadi2016regularization} and segmentation \cite{ji2019invariant}. The idea is that the output of the classification should be the same for the perturbed versions of the same image in the input. For the pixel segmentation, apart from perturbation of the pixel with noise, it should be equivariant under geometrical transformations such as rotation \cite{ji2019invariant}. We consider the rotational equivariance regularization for the last convolutional layer, i.e. the rotation of the image $\alpha$ degrees should produce the same output with $\alpha$ degrees rotation. Consider $T_{\alpha}$ as operator which rotates the image $\alpha$ degree clockwise. We consider the regularization loss
\begin{equation}
 L_{reg}=\sum_{\alpha \in \mathcal{R}} \|{A} (\mathbf{x})-T^{-1}_{\alpha}(T_{\alpha}{{A} (\mathbf{x})}) \|_2
\end{equation}
where $\mathcal{R}$ is a set of degrees of rotation, and ${A} (\mathbf{x})$ is the features of the last convolution layer of the image $\mathbf{x}$. Similar regularization loss has also been used in semi-supervised medical image segmentation \cite{li2020transformation}.

We train the entire network with a weighted average of the cross-entropy label loss and the pixel regularization loss i.e.
\begin{equation}
\label{eq:totoalloss}
L=L_{ce}+\lambda L_{reg}
\end{equation}
where $\lambda$ is a regularization parameter which determines the trade-off between the two losses.

\subsection{Segmentation network}
The previous subsections, we describe the procedure to obtain initial segmentation mask. However, it has been shown that the results can be improved if these masks are used as the ground truth for a semantic segmentation network with pixel supervision \cite{papandreou2015weakly, wei2018revisiting}. We used Deeplab v3 \cite{chen2017rethinking} segmentation network as used in \cite{wei2018revisiting} for the same purpose.  

We use the same loss as in \cite{wei2017object, wei2018revisiting}, which is a weighted average of the cross entropy losses of the initial masks and the currently estimated mask. 

\section{Experimental results}
In this section, we evaluate the performance of our method. We create a dataset by combining RGB images in the Corsican fire dataset \cite{toulouse2017computer}, and non-fire images from the Image-net dataset. We divided the resulting dataset to train/test/evaluation sets by 60, 20, 20 percentages, respectively. Note that the Corsican fire dataset contains pixel-wise segmentation masks, which is only used in the evaluation and test datasets, and only image level labels are used in the training. The parameter $\lambda$ in (\ref{eq:totoalloss}) is set to $.01$. Four values of $\mathcal{R}=\{0,90,180,270\}$ are used as the rotation degrees  in (\ref{eq:totoalloss}).

The proposed regularized CAM network is initialized by the pre-trained Image-net weights with the VGG backbone, and is trained by the ADAM optimizer \cite{kingma2014adam} with initial learning rate of $3 \times 10^{-5}$, and the weight decay of $10^{-6}$. The network is trained for 50 epochs on the dataset. 
The regularization parameter $\lambda$  in \ref{eq:totoalloss} is set to $.6$.  The threshold for obtaining the masks in the CAM method is set to $.45$ and for the mid-level visualization $.55$, to have the best performance on the evaluation set.

The segmentation network, which is a deeplab v3 network \cite{chen2017rethinking}, is trained by the ADAM optimizer \cite{kingma2014adam} with the initial learning rate $5 \times 10^{-5}$, and the weight decay $10^{-6}$. 

\begin{table*}[]
\centering
\begin{tabular}{|l|l|l|l|l|l|}
\hline
Method &CAM \cite{zhou2016learning}             & \cite{wei2018revisiting}                & proposed    \\ \hline
 IOU &58.38                    & 61.26                  & 72.86         \\ \hline
\end{tabular}
\caption{Average IOU on the test set for our proposed method compared to other weakly supervised segmentation methods.}
\label{tab:stateoftheart}
\end{table*} 

\begin{table*}[]
\centering
\begin{tabular}{|l|l|l|l|l|l|}
\hline
        Stage &Mid-level vis.           &  Mid-level vis.+reg. loss           & Mid-level vis.+reg. loss+ segmentation network     \\ \hline
IOU &65.29                     & 67.37                   & 72.86                       \\ \hline
\end{tabular}
\caption{Average IOU on the test set by applying each stage of the proposed method.}
\label{tab:stage}
\end{table*}

\begin{figure*}
\centering
\begin{subfigure}[b]{0.16\textwidth} \includegraphics[width=\textwidth]{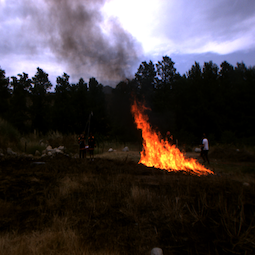} {\\ Original image \\. \\ \hfill} \end{subfigure}
\begin{subfigure}[b]{0.16\textwidth} \includegraphics[width=\textwidth]{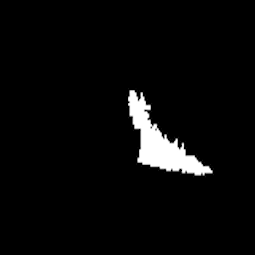}{\\Groundtruth mask \\. \\ \hfill}\end{subfigure}
\begin{subfigure}[b]{0.16\textwidth} \includegraphics[width=\textwidth]{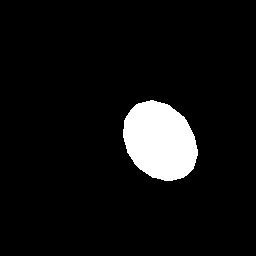}{\\ CAM \cite{zhou2016learning} \\. \\ \hfill}\end{subfigure}
\begin{subfigure}[b]{0.16\textwidth} \includegraphics[width=\textwidth]{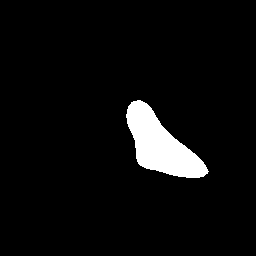} {\\ mid-level visulization mask \\.  \hfill} \end{subfigure}
\begin{subfigure}[b]{0.16\textwidth} \includegraphics[width=\textwidth]{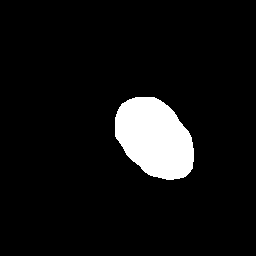} {\\ Method of \cite{wei2018revisiting} \\. \\ \hfill} \end{subfigure}
\begin{subfigure}[b]{0.16\textwidth} \includegraphics[width=\textwidth]{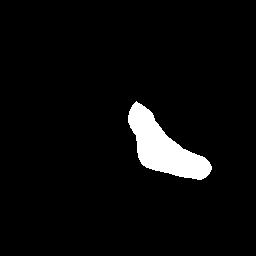} {\\ Proposed  \\. \\ \hfill} \end{subfigure}

\begin{subfigure}[b]{0.16\textwidth} \includegraphics[width=\textwidth]{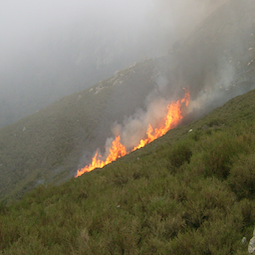} {\\ Original image \\. \\ \hfill} \end{subfigure}
\begin{subfigure}[b]{0.16\textwidth} \includegraphics[width=\textwidth]{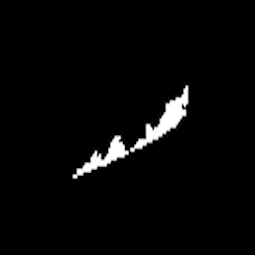}{\\Groundtruth mask \\. \\ \hfill}\end{subfigure}
\begin{subfigure}[b]{0.16\textwidth} \includegraphics[width=\textwidth]{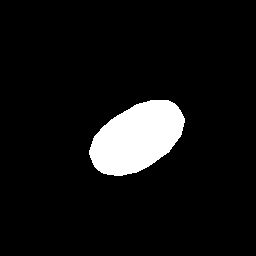}{\\ CAM \cite{zhou2016learning} \\. \\ \hfill}\end{subfigure}
\begin{subfigure}[b]{0.16\textwidth} \includegraphics[width=\textwidth]{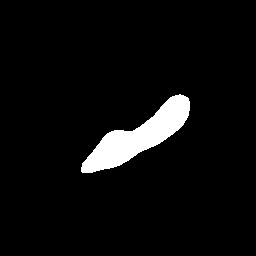} {\\ mid-level visulization mask \\.  \hfill} \end{subfigure}
\begin{subfigure}[b]{0.16\textwidth} \includegraphics[width=\textwidth]{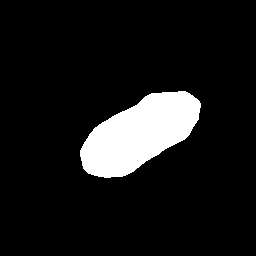} {\\ Method of \cite{wei2018revisiting} \\. \\ \hfill} \end{subfigure}
\begin{subfigure}[b]{0.16\textwidth} \includegraphics[width=\textwidth]{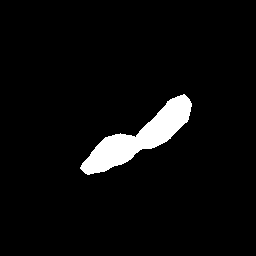} {\\ Proposed \\. \\ \hfill} \end{subfigure}
\caption{Example of the segmented mask obtained by the different stages of the proposed method compared to the other weakly supervised methods}
\label{fig:segmented1}
\end{figure*}

To the best of our knowledge, this paper is the first work that considers weakly supervised  fire segmentation. In order to compare our method, we consider the method in \cite{wei2018revisiting}, which is a multi-class weakly-supervised segmentation method. We changed the output to the binary, to be adapted to our problem, and initialized by the CAM mask. This approach is among the state-of-the-art for weakly supervised image segmentation. Although, the methods proposed for multi-class segmentation may not be optimal for the binary case, since there is no weakly supervised method for fire segmentation, this is the most relevant choice for comparison. We found applying the approach based on dilated convolution instead of CAM as suggested in the paper does not improve the results for the binary fire classification, since as discussed the CAM method does not focus on discriminative parts of the fire.  Table \ref{tab:stateoftheart} reports the performance of the proposed method compared to CAM and the baseline. The thresholding parameter in all methods are set to have optimal performance on the validation set.
Our method outperforms significantly compared to the baseline and CAM. The main reason is due to better initialization segmentation mask by using features in the mid-layers, and applying the equivariance regularization. The effect of each stage in our proposed method (adding the regularized loss, and adding the segmentation network) is compared in table \ref{tab:stage}. As it can be seen, both proposed stages improve the results.  
In Fig. \ref{fig:segmented1}, the resulting masks of our proposed method and other methods are illustrated. As it can be seen the mid-layer visualization captures finer details of the mask.

\section{Conclusions}
In this paper, we proposed a method for fire segmentation while only image labels are available. We found that, unlike general case,  the CAM method, , does not focus on the discriminative part of the fire and covers most part of the object in fire images. We further improve the CAM localization by mid-layer visualization of features in CNN, and adding a regularization loss. Our results show that our method outperforms other methods in the obtaining initial masks and also after training a segmentation network with these initial masks as ground truth.

 \bibliography{IEEEtranBST2/ref} 
\bibliographystyle{IEEEtranBST2/IEEEtran}

\end{document}